\documentclass{article}
\usepackage{spconf,amsmath,graphicx,hyperref}
\usepackage{cite}
\usepackage{booktabs}
\usepackage{amsmath}
\usepackage{tikz}
\usepackage{amsfonts}
\usepackage{inconsolata}
\usepackage{times}
\usepackage{latexsym}
\usepackage{listings}
\usepackage{hyperref}
\usepackage[T1]{fontenc}

\usepackage[utf8]{inputenc}
\usepackage{array} 

\newcolumntype{P}[1]{>{\centering\arraybackslash}p{#1}}  
\newcolumntype{G}[1]{>{\raggedleft\arraybackslash}p{#1}} 
\usepackage{microtype}
\usepackage{multirow}
\usepackage{inconsolata}
\usepackage{amsmath}
\usepackage{graphicx}
\usepackage{microtype}
\usepackage{tcolorbox}
\tcbuselibrary{listingsutf8}
\newtcolorbox{promptbox}{
    colback=gray!5,
    colframe=gray!40,
    fontupper=\small\ttfamily,
    width=\linewidth,
    boxrule=0.5pt,
    breakable,
    listing only,
    listing options={basicstyle=\small\ttfamily, breaklines=true}
}
\usepackage{colortbl}
\usepackage{xcolor}
\definecolor{lightgray}{gray}{0.9}
\usepackage{booktabs}
\usepackage{color, soul}
\usetikzlibrary{arrows.meta,positioning}

\newcommand{\gcol}[1]{\textcolor{green!50!black}{\textbf{(#1)}}}

\title{Plug-and-Play Emotion Graphs for Compositional Prompting in Zero-Shot Speech Emotion Recognition}
\name{Jiacheng Shi$^{\star}$ \qquad Hongfei Du$^{\star}$ \qquad Y. Alicia Hong$^{\ddagger
}$ \qquad Ye Gao$^{\star}$}
\address{$^{\star}$ College of William \& Mary,  $^{\ddagger}$ George Mason University\\
\small \texttt{\{jshi12, hdu02, ygao18\}@wm.edu, yhong22@gmu.edu}}
\begin{document}
%
\maketitle
\begin{abstract}
Large audio-language models (LALMs) exhibit strong zero-shot performance across speech tasks but struggle with speech emotion recognition (SER) due to weak paralinguistic modeling and limited cross-modal reasoning. We propose Compositional Chain-of-Thought Prompting for Emotion Reasoning (CCoT-Emo), a framework that introduces structured Emotion Graphs (EGs) to guide LALMs in emotion inference without fine-tuning. Each EG encodes seven acoustic features (e.g., pitch, speech rate, jitter, shimmer), textual sentiment, keywords, and cross-modal associations. Embedded into prompts, EGs provide interpretable and compositional representations that enhance LALM reasoning. Experiments across SER benchmarks show that CCoT-Emo outperforms prior SOTA and improves accuracy over zero-shot baselines\href{https://github.com/jiachengQAQ/EMO-COT}.
\end{abstract}

\section{Introduction}
\label{sec:intro}
Large audio-language models (LALMs)~\cite{chu2024qwen2,xu2025qwen2,ding2025kimi} have recently demonstrated strong capabilities across diverse speech-oriented tasks, including instruction-following, speech recognition, and audio-based question answering. Among these, speech emotion recognition (SER)~\cite{el2011survey, lin2024emo,lin2025improving} is critical for enabling emotionally intelligent human-computer interactions. Despite promising results, LALMs often focus on linguistic semantics and underrepresent paralinguistic cues such as pitch, speech rate, volume, jitter, shimmer, intensity, and articulation rate, all of which are crucial for emotional 
understanding~\cite{wang2024blsp}.


To address these limitations, recent efforts have integrated Whisper-based~\cite{radford2023robust} acoustic encoders with instruction-tuned language models. These include approaches that align ASR corpora and fine-tune models for emotion-aware generation~\cite{wang2024blsp}, Whisper-BEATs dual-encoders with Q-Former and LoRA-adapted backbones for multi-task instruction tuning~\cite{tang2023salmonn}, and joint ASR-emotion models under low-resource constraints~\cite{geng2025osum}. While effective, such methods rely on annotated corpora and task-specific training, which limits generalization.



To reduce reliance on fine-tuning and further improve models’ understanding of emotional cues, prompting-based methods have emerged, leveraging chain-of-thought (CoT) reasoning to guide LALMs through explicit, step-by-step emotional inference~\cite{murzaku2025omnivox,zhao2025steering}. These approaches aim to surface latent paralinguistic reasoning by prompting LALMs to interpret acoustic signals through intermediate, natural language explanations. However, due to their unstructured nature, CoT prompts are vulnerable to hallucination and error propagation. Moreover, their interpretability remains largely implicit, offering limited transparency into how emotional cues are derived from complex multimodal signals. 

Inspired by recent advances in CoT prompting within the vision-language domain~\cite{mitra2024compositional}, which leverage structured representations of objects, attributes, and their relations to enable modular reasoning, we develop a zero-shot CoT approach that employs structured graph representations to enhance cross-modal reasoning and interpretability in audio-language emotion understanding.We introduce Compositional Chain-of-Thought Prompting for Emotion Reasoning (CCoT-Emo), a zero-shot prompting framework that equips large audio-language models (LALMs) with structured intermediate reasoning. CCoT-Emo constructs an Emotion Graph from input audio and transcripts by extracting seven acoustic features \textbf{(e.g., pitch, speech rate, volume, jitter, shimmer, intensity, and articulation rate), textual sentiment, keywords, and their cross-modal associations}. This graph is embedded into the prompt to serve as a compositional and interpretable reasoning trace. By replacing unstructured CoT reasoning with a symbolic graph, CCoT-Emo improves interpretability, reduces reliance on fine-tuning, and generalizes effectively across diverse speech emotion recognition benchmarks.

\begin{figure*}[t]
    \vspace{-10mm}
    \centering
    \includegraphics[width=0.8\textwidth]{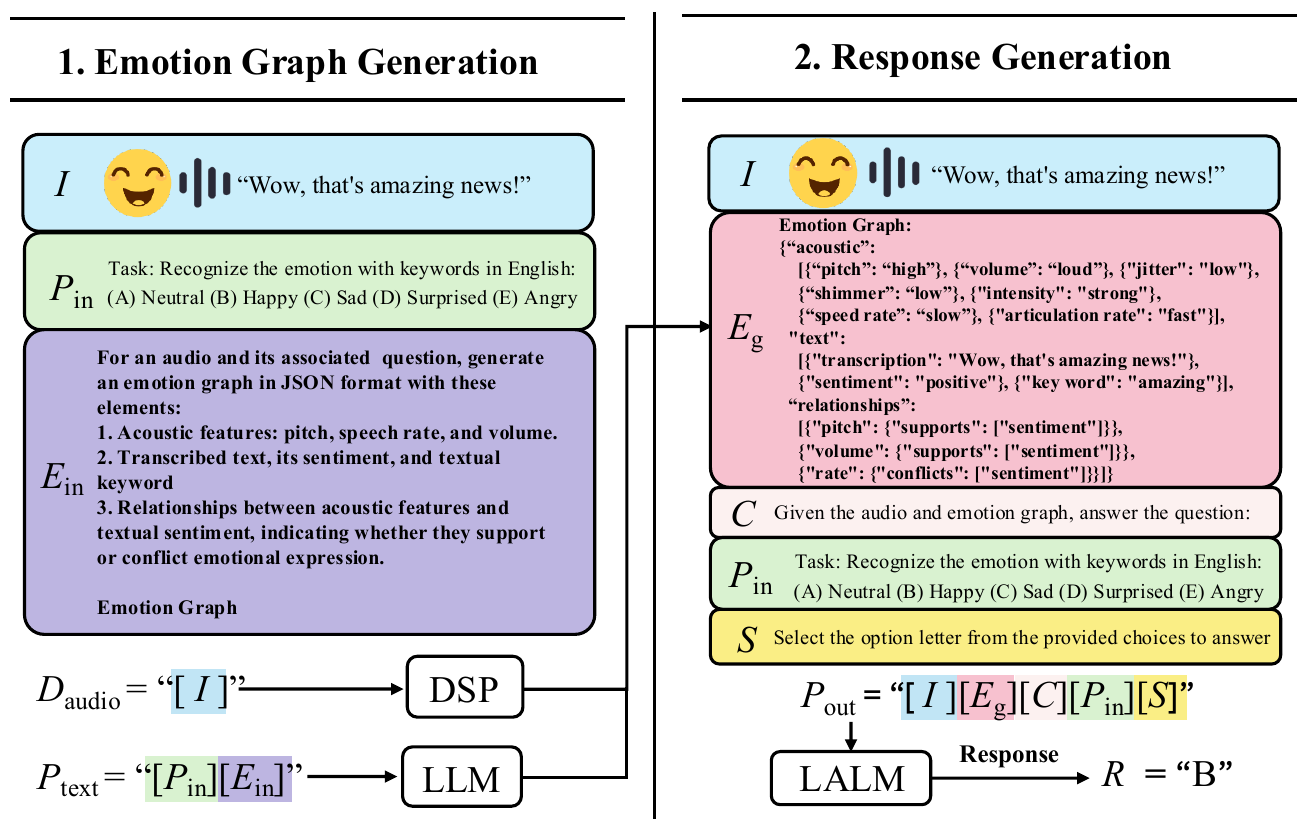}
    \caption{Overview of full CCoT-Emo. Stage 1 builds emotion graph (EG) by extracting acoustic features via digital signal processing (DSP)~\cite{oppenheim1999discrete} and deriving textual sentiment, keywords, and cross-modal relationships using an LLM. Stage 2 prompts the LALM with the audio, EG, and task instruction to predict the emotion label. Components unique to CCoT-Emo are in \textbf{bold}.}
    \label{fig:main}
    \vspace{-2mm}
\end{figure*}

Our contributions are as follows:  
(1) We introduce Compositional Chain-of-Thought for Emotion Reasoning (CCoT-Emo), a zero-shot prompting framework that leverages structured Emotion Graphs to enhance cross-modal compositional reasoning in large audio-language models. (2) We present CCoT-Emo as a plug-and-play, fine-tuning-free approach that improves interpretability and remains broadly compatible with diverse LALM architectures with minimal adaptation effort.
(3) We conduct extensive evaluations showing that CCoT-Emo improves SER performance across multiple popular LALMs, achieving gains of 9.1\%, 8.3\%, and 7.2\% on Qwen2-Audio, Qwen2.5-Omni, and Kimi-Audio, respectively. It also outperforms the prior state-of-the-art method by 3.7\% on average.

\section{Method}
\textbf{Preliminaries.} 
We study speech emotion recognition (SER) with large audio-language models (LALMs) that perform emotion reasoning over acoustic and textual inputs via language modeling. Given an audio input $I$ and a prompt $P_{in}$, the audio is encoded by a speech encoder $\psi_{\phi}(\cdot)$ and projected to discrete token-like embeddings via $\tau(\cdot)$, while the text is tokenized by $l(\cdot)$. The audio and textual inputs are encoded and fused into a unified representation, which is then fed into the language model $f_{\theta}(\cdot)$ to generate the response $R$:
\begin{align}
    R = f_{\theta}(\tau(\psi_{\phi}(I)),\ l(P_{in}))
\end{align}
This token-level interface is common across instruction-tuned LALMs, regardless of specific choices for $\psi_{\phi}$ or $f_{\theta}$.

\noindent \textbf{Step 1: Emotion Graph Generation.} The first stage of our framework constructs an intermediate Emotion Graph $E_g$, which explicitly represents the interaction between acoustic and textual emotional cues. This graph is built without task-specific supervision and serves as a structured reasoning trace for downstream emotion inference.

Seven acoustic features including pitch, speech rate, volume, jitter, shimmer, intensity, and articulation rate are extracted using standard digital signal processing (DSP) techniques~\cite{oppenheim1999discrete}. Specifically, we employ the openSMILE toolkit~\cite{opensmile2010} to compute each feature over the full utterance. Each feature is then discretized into categorical labels: low, normal, or high, based on empirical value distributions computed over the corpus. These discrete labels serve as interpretable attribute tokens within the Emotion Graph, contributing diverse paralinguistic perspectives across prosodic variation, voice quality, and temporal dynamics.

Textual features are extracted from the utterance transcription using a RoBERTa-based sentiment classifier, which predicts sentence-level sentiment polarity as positive, negative, or neutral. Emotionally salient keywords are additionally identified using KeyBERT~\cite{grootendorst2020keybert}, a semantic similarity-based extractor. These high-level textual representations support the construction of cross-modal relationships in the $E_g$.

Cross-modal relationships are inferred using GPT-4. For example, the model is prompted with: \textit{“Given acoustic cues (e.g., pitch: high, volume: loud, etc.), determine whether each one supports, contradicts, or is neutral with respect to the sentiment: positive.”} This enables the model to evaluate whether each acoustic cue reinforces or contradicts the emotional content of text, capturing nuanced multimodal interactions.

The final Emotion Graph $E_g$ integrates acoustic features from $D_{audio}$, textual sentiment and keywords from $P_{text}$, and their cross-modal relationships. All components are computed independently and serialized into a unified JSON structure, as shown in red under Response Generation in Fig.~\ref{fig:main}. This structured representation provides a compositional and interpretable foundation for downstream emotional reasoning.

\begin{table*}[htp]
    \centering
    \vspace{-12mm}
    \scriptsize
    \begin{tabular}{lcccccc}
    \toprule
    
    \multicolumn{7}{c}{\textbf{Speech Emotion Recognition (Acc\%)}} \\
    \midrule
    \textbf{Method} & \textbf{IEMOCAP} & \textbf{MELD} & \textbf{ESD} & \textbf{MER test1} & \textbf{MER test2} & \textbf{Overall} \\
    \midrule
    \multicolumn{7}{l}{\textbf{\emph{Encoder-based Classification Models}}} \\
    \midrule
    HuBERT-Large~\cite{hsu2021hubert} & 64.6 & 53.2 & 70.5 & 55.6 & 45.3 & 57.8 \\
    wav2vec2-Large~\cite{baevski2020wav2vec} & 69.3 & 54.8 & 64.0 & 41.2 & 40.6 & 54.0 \\
    WavLM-Large~\cite{chen2022wavlm} & 68.9 & 54.6 & 70.3 & 48.3 & 42.8 & 57.0 \\
    CLEP-DG~\cite{shi2025clep} & 74.7 & 55.2 & 69.4 & 54.8 & 49.5 & 60.7 \\
    \midrule
    \multicolumn{7}{l}{\textbf{\emph{LLM-based Generative Models}}} \\
    \midrule
    Text+LLM & 54.8 & 54.0 & 54.3 & 47.2 & 44.5 & 50.9 \\
    Whisper+LLM~\cite{radford2023robust} & 57.1 & 53.8 & 53.7 & 49.4 & 46.9 & 52.2 \\
    OSUM~\cite{geng2025osum} & 52.8 & 53.1 & 50.9 & 62.5 & \underline{55.2} & 54.9 \\
    SALMONN-7B~\cite{tang2023salmonn} & 67.0 & 32.9 & 51.8 & 45.8 & 41.7 & 47.8 \\
    SenseVoice-L~\cite{an2024funaudiollm} & 71.3 & 54.7 & 65.6 & 59.3 & 56.7 & 61.5 \\
    BLSP-Emo~\cite{wang2024blsp} & \underline{76.1} & 57.2 & 72.2 & 60.0 & 54.7 & 64.0 \\
    \midrule
    Qwen2-Audio~\cite{chu2024qwen2} & 65.5 & 55.5 & 57.1 & 52.9 & 47.4 & 55.8 \\
    Qwen2-Audio-ZS-CoT & 66.7 & 56.3 & 54.4 & 49.7 & 42.8 & 54.0 \\
    \textbf{Qwen2-Audio-CCoT-Emo} & 72.7\gcol{+7.2} & 61.3\gcol{+6.2} & 71.6\gcol{+14.5} & \underline{62.7}\gcol{+9.8} & 53.1\gcol{+6.7} & 64.9\gcol{+9.1} \\
    \midrule
    Qwen2.5-Omni~\cite{xu2025qwen2} & 64.7 & 57.0 & 59.3 & 54.3 & 50.1 & 57.0 \\
    Qwen2.5-ZS-CoT & 65.0 & 57.9 & 55.6 & 47.9 & 45.4 & 54.4 \\
    \textbf{Qwen2.5-CCoT-Emo} & 71.2\gcol{+6.5} & \underline{63.4}\gcol{+6.4} & \underline{74.2}\gcol{+12.3} & 61.6\gcol{+7.3} & 54.9\gcol{+4.8} & \underline{65.3}\gcol{+8.3} \\
    \midrule
    Kimi-Audio~\cite{ding2025kimi} & 71.5 & 59.1 & 69.5 & 53.7 & 48.9 & 60.5 \\
    Kimi-Audio-ZS-CoT & 69.4 & 60.8 & 67.6 & 47.1 & 46.6 & 58.3 \\
    \textbf{Kimi-Audio-CCoT-Emo} & \textbf{78.4}\gcol{+6.9} & \textbf{64.9}\gcol{+5.5} & \textbf{76.6}\gcol{+7.1} & \textbf{63.4}\gcol{+8.5} & \textbf{56.4}\gcol{+7.5} & \textbf{67.7}\gcol{+7.2} \\
    \bottomrule
    \end{tabular}
    \caption{
Zero-shot speech emotion recognition (SER) accuracy (\%) across five benchmark datasets: IEMOCAP, MELD, ESD, MERBench test1, and MERBench test2. Results are reported for both encoder-based classification models and LLM-based generative models. \textbf{Bold} indicates the best performance on each dataset, while \underline{underline} indicates the second-best result.
}
    \label{tab:ser}
    \vspace{-2mm}
\end{table*}

\noindent\textbf{Step 2: Response Generation.} In the second stage, the emotion graph $E_g$ serves as a structured intermediate representation for zero-shot emotion prediction. The model receives the audio input $I$ and a prompt $P_{out}$ composed of task and reasoning components. This prompt concatenates five elements, as illustrated in Figure~\ref{fig:main}, enabling multimodal emotional inference without parameter updates.
\begin{align}
P_{out} = [I][E_g][C][P_{in}][S]
\end{align}
Here, $I$ is a symbolic reference to the raw audio input, $E_g$ is the structured emotion graph generated in the previous stage, and $C$ is a contextual instruction that guides the model to reason over multimodal content:
\textit{"Use the audio and emotion graph as context and answer the following question."} The task-specific prompt $P_{in}$ specifies the SER objective: \textit{"Task: Recognize the emotion with keywords in English: (A) Neutral (B) Happy (C) Sad (D) Surprised (E) Angry."}
To enforce format consistency, we append an output instruction $S$:
\textit{"Select the option letter from the provided choices to answer."} This explicit formatting constraint reduces output ambiguity and is inspired by prompting strategies in LLaVA~\cite{liu2024improved}, though it remains adaptable to other generation formats. The final response $R$ is generated by the model through joint reasoning over audio and text inputs. The raw audio input is encoded by a speech encoder $\psi_{\phi}(I)$ and projected into discrete token-like representations via $\tau(\cdot)$, while the textual prompt is embedded using a language tokenizer $l(\cdot)$:
\begin{align}
R = f_{\theta}(\tau(\psi_{\phi}(I)),\ l(P_{out}))
\end{align}

\section{Experiments}
\textbf{Datasets and Models.} We evaluate CCoT-Emo on four benchmarks: IEMOCAP, MELD, ESD, and MERBench.  
IEMOCAP contains 5,531 English utterances from five dyadic sessions; we follow~\cite{wang2024blsp} in merging \textit{excited} and \textit{happy}, using 5-fold leave-one-session-out cross-validation.  
MELD comprises 8,851 English utterances from the \textit{Friends} TV series labeled with seven emotions; we follow~\cite{wang2024blsp} in removing \textit{disgust} and \textit{fear}.  
ESD offers 33,443 utterances in English and Chinese across five emotions and diverse speakers.  
MERBench test1 and test2 are small Chinese subsets (~350 utterances each) from movies and TV; we exclude the \textit{worried} label following~\cite{wang2024blsp}.  
As ESD and MERBench lack transcripts, we use Whisper for automatic transcription.  
Since CCoT-Emo is training-free, evaluations are conducted directly on test sets.  
We apply our method to Qwen2-Audio~\cite{chu2024qwen2}, Qwen2.5-Omni~\cite{xu2025qwen2}, and Kimi-Audio~\cite{ding2025kimi}.

\begin{table*}[t]

\scriptsize
\vspace{-8mm}

\begin{center}
\resizebox{0.9\textwidth}{!}{
\begin{tabular}{m{0.25\textwidth}P{0.08\textwidth}P{0.08\textwidth}P{0.08\textwidth}P{0.12\textwidth}P{0.12\textwidth}P{0.07\textwidth}}
    \toprule
    \textbf{Model}  & \textbf{IEMOCAP} & \textbf{MELD} & \textbf{ESD} & \textbf{MERBench test1} & \textbf{MERBench test2} & \textbf{Overall} \\
    \midrule
    \textbf{Qwen2.5-Omni-7B-CCoT-Emo} & 71.2 & 63.4 & 74.2 & 61.6 & 54.9 & 65.3 \\
    Qwen2.5-Omni-7B & 64.7 & 57.0 & 59.3 & 54.3 & 50.1 & 57.0 \\
    \midrule
    \quad w/out Acoustic Attribute & 69.0 & 61.1 & 69.2 & 59.0 & 53.3 & 62.4\\ 
    \quad w/out Text Attribute  & 68.1 & 60.7 & 66.8 & 58.3 & 52.7 & 61.4\\ 
    \quad w/out Cross-Modal Relationship & 70.0 & 61.9 & 71.8 & 60.2 & 53.9 & 63.6\\ 
    \midrule
    \quad w/ spectrograms featrue & 67.6 & 59.8 & 63.5 & 56.1 & 50.4 & 59.4\\ 
    \quad w/ LALMs generate acoustic attribute & 70.2 & 62.3 & 71.5 & 58.7 & 52.3 & 63.1\\ 
    \quad w/out JSON Format & 70.1 & 62.2 & 73.0 & 59.1 & 54.8 & 63.9 \\
    \midrule
    Qwen-2.5-Omni-7B with free-form CoT  & 70.9 & 62.3 & 71.5 & 59.4 & 52.9 & 63.4 \\ 
    Qwen-2.5-Omni-3B  & 55.6 & 45.3 & 47.8 & 45.9 & 37.4 & 46.6\\ 
    Qwen-2.5-Omni-3B-CCoT-Emo & 59.8 & 51.1 & 53.0 & 50.2 & 40.8 & 50.9\\ 
    \midrule
    128 Token Length & 71.5 & 63.6 & 73.8 & 61.0 & 54.5 & 64.9\\
    512 Token Length & 70.7 & 62.6 & 74.0 & 61.2 & 54.3 & 64.6\\
    1024 Token Length & 70.4 & 62.5 & 74.1 & 61.2 & 54.2 & 64.4\\
    \bottomrule
\end{tabular}
}
\caption{SER accuracy (\%) for CCoT-Emo ablations on five benchmarks. Each row reports the effect of removing individual components or modifying prompt/model configurations. \textit{Overall} denotes the average accuracy across datasets.}
\end{center}

\vspace{-4.5mm}
\label{tbl:abl_study}

\end{table*}

\noindent\textbf{Baselines.} 
To evaluate the effectiveness of CCoT-Emo, we compare it with two zero-shot baselines. The first is direct prompting, where each LALM receives only a minimal task instruction without structured cues or intermediate reasoning, reflecting raw zero-shot performance. The second is Chain-of-Thought (CoT) prompting~\cite{kojima2022large}, where the model first generates intermediate emotional reasoning given the audio and a reasoning trigger (e.g., \textit{“Let’s think step by step”}), followed by a second-stage prompt combining the audio, task, and generated rationale for final prediction. We adopt the two-stage prompting setup from LLaVA~\cite{liu2024improved}, which has demonstrated improved consistency in multimodal tasks. These baselines isolate the impact of our graph-based compositional prompting.

\section{Main Results}     
\textbf{Speech Emotion Recognition. }
We evaluate the zero-shot performance of CCoT-Emo on five emotion recognition benchmarks using three LALMs: Qwen2-Audio, Qwen2.5-Omni, and Kimi-Audio. As shown in Table~\ref{tab:ser}, CCoT-Emo consistently outperforms both vanilla prompting and zero-shot Chain-of-Thought (ZS-CoT) baselines across all models. Kimi-Audio with CCoT-Emo achieves the highest average accuracy of 67.7\%, surpassing vanilla prompting by +7.2\% and ZS-CoT by +9.4\%, and improving over the previous state-of-the-art BLSP-Emo by 3.7\%. Notably, on MERBench test1 and test2, it yields +8.5\% and +7.5\% gains, respectively, indicating strong generalization to domain-shifted and long-form speech. Dataset-specific baselines like BLSP-Emo and OSUM remain competitive on IEMOCAP and MERBench, respectively, due to in-domain training and language-specific pretraining. CCoT-Emo also enhances Qwen2-Audio and Qwen2.5-Omni by +6.1\% and +8.3\%, respectively. On ESD, Qwen2-Audio-CCoT-Emo reaches 71.6\%, a +14.5\% absolute improvement, underscoring the benefit of structured audio-text conditioning. In contrast, ZS-CoT underperforms across all settings, likely due to irrelevant intermediate reasoning disrupting emotion inference, particularly in datasets with long-range context. These results demonstrate CCoT-Emo’s effectiveness in improving zero-shot compositional reasoning in LALMs.

\section{Ablation Studies}

\noindent\textbf{JSON structure enhances EG utilization.}
To assess the impact of representation format, we replace the structured JSON-based Emotion Graph with an unstructured variant. This yields a 1.4\% accuracy drop, suggesting that standardized formatting enhances model interpretability and emotional inference.

\noindent\textbf{Replacing EGs with free-form CoT.}
To probe the effect of structural abstraction, we replace Emotion Graphs with free-form natural language descriptions that preserve content but lack compositional structure~\cite{dutta2024think}. This results in a 1.9\% performance drop, highlighting the advantage of structured representations in guiding multimodal emotional reasoning.
\noindent\textbf{Substituting DSP with LALM for acoustic attribute extraction.}
\label{ablation:dsp_to_lalm} We assess whether LALMs can replace traditional digital signal processing (DSP) for acoustic feature extraction by prompting the model to describe pitch, speech rate, and volume. This substitution results in a 2.2\% accuracy drop, as LALM-generated descriptions are often vague (e.g., “The tone is expressive”) and lack concrete prosodic cues. This suggests that high-level semantic pretraining alone is insufficient for capturing low-level acoustic detail, reaffirming the value of deterministic and interpretable DSP features for reliable zero-shot emotion inference.

\noindent\textbf{Component-wise ablation in Emotion Graph.}  
We ablate each component of the emotion graph to quantify its contribution. Removing cross-modal relationships reduces SER accuracy by 1.7\%, suggesting their role in aligning acoustic and textual signals. Excluding acoustic attributes (e.g., pitch, volume, speech rate, etc.) results in a 2.9\% drop, highlighting their prosodic relevance. The largest degradation (3.9\%) occurs when textual attributes (sentiment, keywords) are removed, confirming their importance as direct semantic cues. Replacing DSP-derived symbolic acoustic features with spectrograms yields an additional 5.9\% drop, indicating that continuous and uninterpretable representations lack the structured abstraction required for zero-shot symbolic reasoning.

\noindent\textbf{Effect of LALM size.}
We compare CCoT-Emo on Qwen2.5-Omni 3B and 7B parameters. The 3B variant improves SER accuracy by 3.9\%, while the 7B version achieves an 8.3\% gain. This trend suggests that larger models better exploit structured intermediate representations for emotion reasoning.

\noindent\textbf{Effect of Emotion Graph size.}  
We examine the effect of emotion graph length by varying token limits from 128 to 1024. A 256-token configuration yields the highest accuracy, while both shorter (128: --0.4\%) and longer variants (512: --0.7\%, 1024: --0.9\%) degrade performance. These results indicate that compact, well-balanced emotion graphs offer optimal guidance for zero-shot multimodal reasoning.

\section{Conclusion}
We introduce CCoT-Emo, a structured zero-shot prompting framework that enhances emotion reasoning in LALMs through intermediate Emotion Graphs encoding acoustic, textual, and cross-modal cues. Without fine-tuning or task-specific supervision, CCoT-Emo consistently improves SER performance across five benchmarks. These results underscore the value of symbolic and interpretable representations for guiding multimodal emotional inference in large audio models.

\bibliographystyle{IEEEbib}
\bibliography{strings,main}

@inproceedings{lin2025improving,
  title={Improving speech emotion recognition in under-resourced languages via speech-to-speech translation with bootstrapping data selection},
  author={Lin, Hsi-Che and Lin, Yi-Cheng and Chou, Huang-Cheng and Lee, Hung-yi},
  booktitle={ICASSP 2025-2025 IEEE International Conference on Acoustics, Speech and Signal Processing (ICASSP)},
  pages={1--5},
  year={2025},
  organization={IEEE}
}

@article{hsu2021hubert,
  title={HuBERT: Self-supervised speech representation learning by masked prediction of hidden units},
  author={Hsu, Wei-Ning and et al.},
  journal={IEEE/ACM Transactions on Audio, Speech, and Language Processing},
  year={2021}
}

@article{baevski2020wav2vec,
  title={wav2vec 2.0: A framework for self-supervised learning of speech representations},
  author={Baevski, Alexei and Zhou, Yuhao and Mohamed, Abdelrahman and Auli, Michael},
  journal={Advances in neural information processing systems},
  volume={33},
  pages={12449--12460},
  year={2020}
}

@article{chen2022wavlm,
  title={WavLM: Large-Scale Self-Supervised Pre-Training for Full Stack Speech Processing},
  author={Chen, Shu-wen and others},
  journal={IEEE Journal of Topics in Signal Processing},
  year={2022}
}

@inproceedings{radford2023robust,
  title={Robust speech recognition via large-scale weak supervision},
  author={Radford, Alec and Kim, Jong Wook and Xu, Tao and Brockman, Christine and Sutskever, Ilya},
  booktitle={International conference on machine learning},
  year={2023},
  organization={PMLR}
}

@inproceedings{lin2024emo,
  title={Emo-bias: A Large Scale Evaluation of Social Bias on Speech Emotion Recognition},
  author={Lin, Yi-Cheng and Wu, Haibin and Chou, Huang-Cheng and Lee, Chi-Chun and Lee, Hung-yi},
  booktitle={Proc. Interspeech 2024},
  pages={4633--4637},
  year={2024}
}

@article{wang2024blsp,
  title={BLSP-Emo: Towards Empathetic Large Speech-Language Models},
  author={Wang, Chen and Liao, Minpeng and Huang, Zhongqiang and Wu, Junhong and Zong, Chengqing and Zhang, Jiajun},
  journal={arXiv preprint arXiv:2406.03872},
  year={2024}
}

@article{tang2023salmonn,
  title={Salmonn: Towards generic hearing abilities for large language models},
  author={Tang, Changli and Yu, Wenyi and Sun, Guangzhi and Chen, Xianzhao and Tan, Tian and Li, Wei and Lu, Lu and Ma, Zejun and Zhang, Chao},
  journal={arXiv:2310.13289},
  year={2023}
}

@article{geng2025osum,
  title={OSUM: Advancing Open Speech Understanding Models with Limited Resources in Academia},
  author={Geng, Xuelong and Wei, Kun and Shao, Qijie and Liu, Shuiyun and Lin, Zhennan and Zhao, Zhixian and Li, Guojian and Tian, Wenjie and Chen, Peikun and Li, Yangze and others},
  journal={arXiv preprint arXiv:2501.13306},
  year={2025}
}

@article{chu2024qwen2,
  title={Qwen2-audio technical report},
  author={Chu, Yunfei and Xu, Jin and Yang, Qian and Wei, Haojie and Wei, Xipin and Guo, Zhifang and Leng, Yichong and Lv, Yuanjun and He, Jinzheng and Lin, Junyang and others},
  journal={arXiv preprint arXiv:2407.10759},
  year={2024}
}

@inproceedings{mitra2024compositional,
  title={Compositional chain-of-thought prompting for large multimodal models},
  author={Mitra, Chancharik and et al.},
  booktitle={Proceedings of the IEEE/CVF Conference on Computer Vision and Pattern Recognition},
  year={2024}
}

@article{zhao2025steering,
  title={Steering Language Models to Stable Speech Emotion Recognition via Contextual Perception and Chain-of-Thought},
  author={Zhao, Zhixian and Zhu, Xinfa and Xinsheng and Wang and Xie, Lei and et al.},
  journal={arXiv preprint arXiv:2502.18186},
  year={2025}
}

@article{murzaku2025omnivox,
  title={OmniVox: Zero-Shot Emotion Recognition with Omni-LLMs},
  author={Murzaku, John and Rambow, Owen},
  journal={arXiv preprint arXiv:2503.21480},
  year={2025}
}

@inproceedings{opensmile2010,
  title={Opensmile: The munich versatile and fast open-source audio feature extractor},
  author={Eyben, Florian and W{\"o}llmer, Martin and Schuller, Bj{\"o}rn},
  booktitle={Proceedings of the 18th ACM international conference on Multimedia},
  year={2010}
}

@inproceedings{liu2024improved,
  title={Improved baselines with visual instruction tuning},
  author={Liu, Haotian and Li, Chunyuan and Li, Yuheng and Lee, Yong Jae},
  booktitle={Proceedings of the IEEE/CVF Conference on Computer Vision and Pattern Recognition},
  pages={26296--26306},
  year={2024}
}

@article{kojima2022large,
  title={Large language models are zero-shot reasoners},
  author={Kojima, Takeshi and Gu, Shixiang Shane and Reid, Machel and Matsuo, Yutaka and Iwasawa, Yusuke},
  journal={Advances in neural information processing systems},
  volume={35},
  year={2022}
}

@article{ding2025kimi,
  title={Kimi-Audio Technical Report},
  author={Ding, Ding and Ju, Zeqian and Leng, Yichong and Liu, Songxiang and Liu, Tong and Shang, Zeyu and Shen, Kai and Song, Wei and Tan, Xu and Tang, Heyi and others},
  journal={arXiv preprint arXiv:2504.18425},
  year={2025}
}

@article{el2011survey,
  title={Survey on speech emotion recognition: Features, classification schemes, and databases},
  author={El Ayadi, Moataz and Kamel, Mohamed S and Karray, Fakhri},
  journal={Pattern recognition},
  volume={44},
  number={3},
  pages={572--587},
  year={2011},
  publisher={Elsevier}
}

@article{dutta2024think,
  title={How to think step-by-step: A mechanistic understanding of chain-of-thought reasoning},
  author={Dutta, Subhabrata and Singh, Joykirat and Chakrabarti, Soumen and Chakraborty, Tanmoy},
  journal={arXiv preprint arXiv:2402.18312},
  year={2024}
}

@book{oppenheim1999discrete,
  title={Discrete-time signal processing},
  author={Oppenheim, Alan V},
  year={1999},
  publisher={Pearson Education India}
}

@misc{grootendorst2020keybert,
  author       = {Maarten Grootendorst},
  title        = {KeyBERT: Minimal keyword extraction with BERT.},
  year         = 2020,
  publisher    = {Zenodo},
  version      = {v0.3.0},
  doi          = {10.5281/zenodo.4461265},
  url          = {https://doi.org/10.5281/zenodo.4461265}
}

@article{an2024funaudiollm,
  title={FunAudioLLM: Voice understanding and generation foundation models for natural interaction between humans and LLMs},
  author={An, Keyu and et al.},
  journal={arXiv preprint arXiv:2407.04051},
  year={2024}
}

@article{xu2025qwen2,
  title={Qwen2. 5-omni technical report},
  author={Xu, Jin and Guo, Zhifang and He, Jinzheng and Hu, Hangrui and He, Ting and Bai, Shuai and Chen, Keqin and Wang, Jialin and Fan, Yang and Dang, Kai and others},
  journal={arXiv preprint arXiv:2503.20215},
  year={2025}
}

@inproceedings{shi2025clep,
  title={CLEP-DG: Contrastive Learning for Speech Emotion Domain Generalization via Soft Prompt Tuning},
  author={Shi, Jiacheng and Zhang, Yanfu and Gao, Ye},
  booktitle={Proc. Interspeech 2025},
  pages={4498--4502},
  year={2025}
}

\end{document}